# Complete characterization of a class of privacy-preserving tracking problems

Yulin Zhang and Dylan A. Shell
Department of Computer Science and Engineering
Texas A&M University
yulinzhang, dshell@tamu.edu

*Abstract*—We examine the problem of target tracking whilst simultaneously preserving the target's privacy as epitomized by the robotic panda tracking scenario, which O'Kane introduced at the *2008 Workshop on the Algorithmic Foundations of Robotics* in order to elegantly illustrate the utility of ignorance. The present paper reconsiders his formulation and the tracking strategy he proposed, along with its completeness. We explore how the capabilities of the robot and panda affect the feasibility of tracking with a privacy stipulation, uncovering intrinsic limits, no matter the strategy employed. This paper begins with a one-dimensional setting and, putting the trivially infeasible problems aside, analyzes the strategy space as a function of problem parameters. We show that it is not possible to actively track the target as well as protect its privacy for every nontrivial pair of tracking and privacy stipulations. Secondly, feasibility can be sensitive, in several cases, to the information available to the robot initially. Quite naturally in the one-dimensional model, one may quantify sensing power by the number of perceptual (or output) classes available to the robot. The robot's power to achieve privacy-preserving tracking is bounded, converging asymptotically with increasing sensing power. We analyze the entire space of possible tracking problems, characterizing every instance as either achievable, constructively by giving a policy where one exists (some of which depend on the initial information), or proving the instance impossible. Finally, to relate some of the impossibility results in one dimension to their higher-dimensional counterparts, including the planar panda tracking problem studied by O'Kane, we establish a connection between tracking dimensionality and the sensing power of a one-dimensional robot.

## I. INTRODUCTION

Most roboticists see uncertainty as a problem, something which should be minimized or, if possible, eliminated. As robots become widespread, there will likely be a shift in thinking since robots that know too much can be problematic in their own way. A robot operating in your home, benignly monitoring your activities and your daily routine, for example to schedule vacuuming at unobtrusive times, possesses information that is valuable. There are certainly those who could profit from it. A tension exists between the information necessary for a robot to be useful and information which could be sensitive if mistakenly disclosed or stolen.

The present paper's focus is on the problem of tracking a target whilst preserving the target's privacy. Though fairly narrow, this is a crisply formulated instance of the broader dilemma of balancing the information a robot possesses: the robot must maintain some estimate of the target's pose but information that is too precise is an unwanted intrusion and potential hazard if leaked. The setting we examine, the *panda tracker problem*, is due to O'Kane who expressed the idea of uncertainty being valuable, aloofness having utility. The following, quoted verbatim from O'Kane (2008, p. 1), describes the scenario:

> "A giant panda moves unpredictably through a wilderness preserve. A mobile robot tracks the panda's movements, periodically sensing partial information about the panda's whereabouts and transmitting its findings to a central base station. At the same time, poachers attempt to exploit the presence of the tracking robot—either by eavesdropping on its communications or by directly compromising the robot itself—to locate the panda. We assume, in the worst case, that the poachers have access to any information collected by the tracking robot, but they cannot control its motions. The problem is to design the tracking robot so that the base station can record coarse-grained information about the panda's movements, without allowing the poachers to obtain the fine-grained position information they need to harm the panda."

Note that it is not sufficient for the robot to simply forget or to degrade sensor data via post-processing because the adversary may have compromised these operations, possibly writing the information to separate storage.

One can view the informational constraints as bounds:
1) A maximal- or upper-bound specifies how coarse the tracking information can be. The robot is not helpful in assuring the panda's well-being when this bound is exceeded.
2) A second constraint, a lower-bound, stipulates that if the information is more fine-grained than some threshold, a poacher may succeed in having his wicked way.

The problem is clearly infeasible when the lower-bound exceeds the upper-bound, but what of other circumstances? Is it always possible to ensure that one will satisfy both bounds indefinitely? In the original paper, O'Kane (2008) proposed a tracking strategy for a robot equipped with a two-bit quadrant sensor, showing its successful operation in several circumstances. As no claim of completeness was made, one might well ask: will his strategy work for all feasible bounds? And how are the strategies affected by improvements in the sensing capabilities of the robot? These are the class of questions that are of interest to us.

The last of the preceding questions suggests another, more fundamental, one worth asking—what sensors are appropriate for a given problem? The problem of establishing the minimal information required to perform a particular task and the problem of analyzing the trade-off between information and performance, despite both being fundamental challenges, remain poorly understood territory—the efforts of Donald (1995), Erdmann and Mason (1988), and Mason (1993) notwithstanding. As a consequence, we do not attempt to answer this very general question. However, recognising that little has been said regarding scenarios like the one we study, where too much information can be detrimental, we are able to shed some light on the complexity of the sensors involved by examining the relationship played by our sensor's preimages. While a several authors have examined sensor power from a preimage perspective before (Erdmann, 1995; LaValle, 2011), there are two aspects which are novel in problem we study: (i) Our abstract sensor is parameterized, allowing one to change the number of output classes, and we have a situation where taking the limit of increasing power is instructive (see Theorem 2 and Corollary 1) (ii) We relate abstract sensors for scenarios with different dimensionalities and different numbers of preimages (see Theorem 3 and comment thereafter).

Narrowing attention toward privacy and the particularities of the problems dealt with in this article, the next section details relationships to existing research.

## II. Related work

Information processing is a critical part of what most robots do: without estimating properties of the world, their usefulness will often be hampered, sometimes severely so. Many techniques have been developed in order to maximize the information available to a robot (the range spanning from well-established approaches, e.g., Bourgault et al. (2002), through to recent and ongoing work, e.g., Miller et al. (2016)). But increasingly, people are realizing that granting our computational devices access to too much information involves some risk. When that risk is too great, it undermines the entire technology because ultimately users may balk and instead choose to forgo the benefits of efficiency or convenience. Multiple models have been proposed to think about this tension for data processing more generally, *cf.* formulations in Sweeney (2002) and Dwork (2008). Existing work has also explored privacy for networks, along with routing protocols proposed to increase anonymity (Clarke et al., 2001), and includes wireless sensor problems where distinct notions of spatial and temporal privacy (Kamat et al., 2005, 2009) have been identified.

Some recent work has begun to investigate privacy in settings more directly applicable to robots. Specifically three lines of work propose techniques to help automate the development of controllers that respect some limits on the information they divulge. O'Kane and Shell (2015) formulated a version of the problem in the setting of combinatorial filters where the designer provides additional stipulations describing which pieces of information should always be plausibly indistinguishable and which must never be conflated. The paper describes hardness results and an algorithm for ascertaining whether a design satisfying such a stipulation is feasible. This determination of feasibility for a given set of privacy and utility choices is close in spirit to what this paper has explored for the particular case of privacy preserving tracking: here those choices become quantities $r_p$ and $r_t$. The second important line is that of Wu et al. who explored how to disguise some behavior in their control system's plant model, and showed how to protect this secret behavior by obfuscating the system outputs (Wu and Lafortune, 2014). More recent work expresses both utility and privacy requirements (in a finite automata framework) and proposes algorithms to synthesize controllers satisfying these two constraints (Wu et al., 2016). In the third line, Prorok and Kumar (2016) adopt the differential privacy model to characterize the privacy of heterogeneous robot swarms, so that any individual robot cannot be determined to be of a particular type from macroscopic observations of the swarm.

Additionally, we note that while we have considered panda tracking as a cute realization of this broader informationally constrained problem, wildlife monitoring and protection is an area in which serious prior robotics work exists (e.g., see Bhadauria et al. (2010)) and for which there is substantial and growing interest (Fang et al., 2015).

Finally, we point out that this article is an extended version of our earlier conference paper in Zhang and Shell (2016). That more preliminary work had only a partial characterization of the set of problems as, at that point, we had not yet arrived at solutions for all problem instances with dependencies on the robot's initial information. Thus, while this paper can claim a complete characterization of the problems, the earlier work did not.

## III. Formulating the one-dimensional panda tracking problem

The original problem was posed in two dimensions: the robot and panda, inhabiting a plane that is assumed to be free of obstacles, both move in discrete time-steps, interleaving their motions. A powerful adversary, who is interested in computing the possible locations of the panda, is assumed to have access to the full history of information. Any information is presumed to be used optimally by the adversary in reconstruction of possible locations of the panda—by which we mean that the region that results is both sound (that is, consistently accounted for by the information) but is also tight (no larger than necessary). The problem is formulated without needing to appeal to probabilities by considering only worst-case reasoning and by using motion- and sensor-models characterized by regions and applying geometric operations.

*Information stipulation:* The tracking and privacy requirements were specified as two disks. The robot is constrained to ensure that the region describing possible locations of the panda always fits inside the *tracking disk*, which has the larger diameter of the two. The *privacy disk* imposes the requirement that it always be possible to place the smaller disk wholly inside the set of possible locations.

*Sensor model:* As reflected in the title of his paper, O'Kane considered an unconventional sensor that consists of four

IR sensors*, and outputs only two bits of information per measurement. With origin centered on the robot and an axis parallel to the robot's heading, the sensor outputs the quadrant containing the panda.

*Target motion model:* The panda moves unpredictably with bounded velocity. After a time-step has elapsed, the set of newly feasible locations for the panda is obtained by convolving a disk with the previous time-step's region. The disk must be sized appropriately for the time-step interval and the target's velocity.

Now, by way of simplification, consider pandas and robots that inhabit obstacle-free one-dimensional worlds, each only moving left or right along a line.

*Information stipulation:* Using the obvious one-dimensional analogue, now the robot tracker has to bound its belief about the panda's potential locations to an interval of minimum size $r_p$ ($p$ for privacy) and maximum size $r_t$ ($t$ for tracking).

*Sensor model:* Most simply, the quadrant sensor corresponds to a one-bit sensor indicating whether the panda is on the robot's left- or right-hand side. When the robot is at $u_1$, the sensor indicates whether the panda is within $(-\infty, u_1]$ or $(u_1, \infty)$.

In what follows, we will also explore how modifying the robot's sensing capabilities alters its possible behavior. Thus, we give the robot $c$ set-points $u_1 < u_2 < \cdots < u_c$, each within the robot's control, so that it can determine which of the $c+1$ non-overlapping intervals contains the panda. With $c$ set-points one can model any sensor that fully divides the state space and produces at most $c+1$ observations. Note that this is not a model of any physical sensor of which we are aware. The reader, finding this too contrived, may find later (e.g., for $n$-dimensional tracking, see Lemma 13) that the choice has merit. The case with $c=1$ is the straightforward analogue of the quadrant sensor. Increasing $c$ yields a robot with greater sensing power, since the robot has a wider choice of how observations should inform itself of the panda's location.

*Target motion model:* The convolution becomes a trivial operation on intervals.

Figure 1 provides a visual example with $c=2$. The panda is sensed by the robot as falling within one of the following intervals: $(-\infty, u_1]$, $(u_1, u_2]$, $(u_2, \infty)$, where $u_1, u_2 \in \mathbb{R}$ and $u_1 < u_2$. These three intervals are represented by observation values: 0, 1 and 2. For simplicity, no constraints are imposed on the robot's motion and we assume that at each time-step, the robot can pick positions of $u_1 < \cdots < u_c$ as it likes.

*Notation and model:*

In the 1-dim. problem the panda's location is represented as a single coordinate indexed by discrete time. At stage $k$ the location of the panda is $x_k \in \mathbb{R}$. The robot (and adversary)

*For the reader especially interested in O'Kane's physical robot implementation: he describes tracked item (i.e., the panda) carrying four beacons arranged to give 360° coverage, and four infrared sensors, clustered in the center of the robot, angled so that each senses a quadrant.

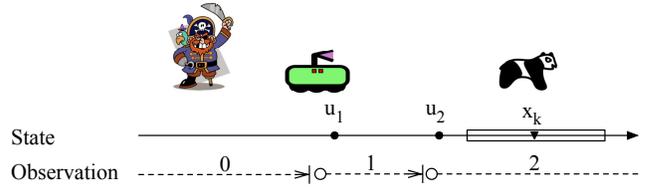

Fig. 1: Panda tracking in one dimension. Inimitable artwork for the robot and panda is adapted from the original in O'Kane (2008, p. 2).

maintains knowledge of the panda's possible location after it moves and between sensing steps by fusing the knowledge accumulated about the panda's possible location, sensor readings, and the movement model. The set of conceivable locations of the panda (a subset of $\mathbb{R}$) is a geometric realization of an information-state or I-state, as formalized and treated in detail by LaValle (2010). In this paper, we take the I-state as an interval.

The movement of the panda per time-step is bounded by length $\frac{\delta}{2}$, meaning that the panda can move at most $\frac{\delta}{2}$ in either direction. We use $\eta_k$ to denote the robot's knowledge of the panda after the observation taken at time $k$. In evolving from $\eta_k$ to $\eta_{k+1}$ the robot's I-state first transits to an intermediate I-state, which we write $\eta_{k+1}^-$, representing the state after adding the uncertainty arising from the panda's movement but before observation $k+1$. Since this update occurs before the sensing information is incorporated, we refer to $\eta_{k+1}^-$ as the *prior* I-state for time $k+1$. Updating I-state $\eta_k$ involves mapping every $x_k \in \eta_k$ to $[x_k - \frac{\delta}{2}, x_k + \frac{\delta}{2}]$, the resultant I-state, $\eta_{k+1}^-$, being the union of the results.

Sensor reading updates to the I-state depend on the values of $u_1(k), u_2(k), \ldots, u_c(k)$, which are under the control of the robot. The sensor reports the panda's location to within one of the $c+1$ non-empty intervals: $(-\infty, u_1(k)], (u_1(k), u_2(k)], (u_2(k), u_3(k)], \ldots, (u_c(k), \infty)$. If we represent the observation at time $k$ as a non-empty interval $y(k)$ then the *posterior* I-state $\eta_k$ is updated as $\eta_k = \eta_k^- \cap y(k)$.

For every stage $k$ the robot chooses a sensing vector $\mathbf{v_k} = [u_1(k), u_2(k), \ldots, u_c(k)]$, $u_i(k) < u_j(k)$ if $i < j$, $u_i(k) \in \mathbb{R}$, so as to achieve the following conditions:

1) *Privacy Preserving Condition (PPC)*: The size of any I-state $\eta_k = [a, b]$ should be at least $r_p$. That is, for every stage $k$, $|\eta_k| = b - a \geq r_p$.
2) *Target Tracking Condition (TTC)*: The size of any I-state $\eta_k = [a, b]$ should be at most $r_t$. That is, for every stage $k$, $|\eta_k| = b - a \leq r_t$.

IV. PRIVACY-PRESERVING TRACKING

Given specific problem parameters, we are interested in whether there is always some $\mathbf{v_k}$ that a robot can select to track the panda while satisfying the preceding conditions.

**Definition 1.** A 1-dim. panda tracking problem is a tuple $P_1 = (\eta_0, r_p, r_t, \delta, c)$, in which

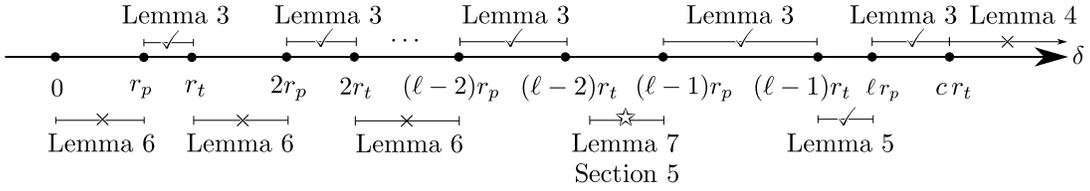

| Name | Condition | Regions |
|---|---|---|
| Teeth | $\delta \in [a\,r_p, a\,r_t]$ | |
| Gaps | $\delta \in (a\,r_t - r_t, a\,r_p)$ | |
| Penultimate | $\delta \in [ar_p + r_p - r_t,\ ar_t - r_p]$ | |
| Filling | $\delta \in (a\,r_t - r_t, a\,r_p)$ and $a\,r_t \geqslant a\,r_p + r_p$ | |
| Beyond control | $\delta \in (c\,r_t, +\infty)$ | |

× means there are no **PP** & **TT** policies

✓ means there are **PP** and **TT** policies

☆ denotes boundary problems with sensitivity to initial conditions

Fig. 2: Roadmap of the results for 1-dim. panda tracking with $c$ sensing parameters.

1) the initial I-state $\eta_0$ describes all the possible initial locations of the panda;
2) the privacy bound $r_p$ gives a lower bound on the I-state size;
3) the tracking bound $r_t$ gives a upper bound on the I-state size;
4) parameter $\delta$ describes the panda's (fastest) motion; and
5) the sensor capabilities are given by the number $c$.

**Definition 2.** The 1-dim. panda tracking problem $P_1 = (\eta_0, r_p, r_t, \delta, c)$ is privacy preservable, written as predicate **PP**$(P_1)$, if starting with $|\eta_0| \in [r_p, r_t]$, there exists some strategy $\pi$ to determine a $\mathbf{v_k}$ at each time-step, such that the Privacy Preserving Condition holds forever. Otherwise, the problem $P_1$ is not privacy preservable: $\neg$ **PP**$(P_1)$.

**Definition 3.** The 1-dim. panda tracking problem $P_1 = (\eta_0, r_p, r_t, \delta, c)$ is target trackable, **TT**$(P_1)$, if starting with $|\eta_0| \in [r_p, r_t]$, there exists some strategy $\pi$ to determine a $\mathbf{v_k}$ at each time-step, such that the Target Tracking Condition holds forever. Otherwise, the problem $P_1$ is not target trackable: $\neg$ **TT**$(P_1)$.

To save space, we say a problem $P_1$ and also its strategy $\pi$ are **PP** if **PP**$(P_1)$. Similarly, both $P_1$ and its strategy $\pi$ will be called **TT** if **TT**$(P_1)$. Putting aside trivially infeasible $P_1$ where $r_p > r_t$, we wish to know which problems are both **PP** and **TT**. Next, we explore the parameter space to classify the various classes of problem instances by investigating the existence of strategies.

*A. Roadmap of technical results*

The results follow from several lemmas. The roadmap in Figure 2 provides a sense of how the pieces fit together to help the reader keep track of the broader picture.

Our approach begins, firstly, by identifying particular actions which we call *basis actions*: one increases the size of the I-state and the other does the opposite. We show that any question about the existence of a strategy can be framed in terms of sequences comprised of these two actions. Next, we divide the space of strategies into 'teeth' and 'gaps' according to the panda's speed. The teeth regions and the last gap region describe under-constrained problems that have **PP** and **TT** tracking strategies for all initial I-states that satisfy $|\eta_0| \in [r_p, r_t]$. The region with uncertainty that grows so quickly as to be beyond control and the remaining gaps (except the penultimate one) are over-constrained problems that have no **PP** and **TT** tracking strategies for any initial I-states satisfying $|\eta_0| \in [r_p, r_t]$. The second-to-last gap describes boundary problems that transition between the under-constrained problems and over-constrained ones. To proceed further, one needs to consider circumstances where the existence of a suitable strategy depends on the initial information available to the robot. These are a logically distinct class of solution and are, therefore, presented in a major section of their own (Section V). Before that segue, a summary of the results established up to that point, including a visual representation of the parameter space, is provided; it may aid the reader to glance ahead to Section IV-D and Figure 6.

*B. Basis actions for privacy-preserving tracking*

A challenge in dealing with 1-dim. panda tracking is the fact that the space of sensor configurations (i.e., the choices of $\mathbf{v_k}$) is continuous, making it infeasible to search through all choices. To resolve this issue, consider those sensor configurations which split the prior I-states into parts of equal size. Thinking in worst-case terms, a suicidal panda would choose the least convenient place to move to. Thus, compared with unevenly partitioned configurations, evenly partitioned configurations have both a "finer" largest I-state and a "safer" smallest I-state. Moreover, evenly sized splits mean that the size of the I-state after the observation is determined and depends only on the number of intervals.

Let $s(i)$ denote the choice of evenly dividing the prior I-state interval into $i$ parts (the mnemonic is $s$ for *split*). Then the following lemma states that it is sufficient to consider strategies consisting solely of $s(i)$ actions to determine whether violation of either the privacy or tracking constraints is inevitable.

**Lemma 1.** *For any problem $P_1 = (\eta_0, r_p, r_t, \delta, c)$, $\mathbf{PP}(P_1) \wedge \mathbf{TT}(P_1)$ if and only if there exists a $\mathbf{PP}$ and $\mathbf{TT}$ strategy that only consists of action $s(i)$, where $i \in \{1, 2, \ldots, c+1\}$.*

*Proof:* $\Leftarrow$: Holds trivially. $\Rightarrow$: For $\mathbf{PP}$ and $\mathbf{TT}$ strategy $\pi$, then we can always construct a $\mathbf{PP}$ and $\mathbf{TT}$ strategy $\pi'$ consisting of actions $s(i)$, where $i \in \{1, 2, \ldots, c+1\}$. Under strategy $\pi$, the sensing vector at stage $k$ is $\mathbf{v_k}$, and the prior I-state is divided into $i_k$ parts. The smallest and largest possible sizes of the resulting I-state are denoted as $\min_{\eta \in \mathbf{v_k}} |\eta|$ and $\max_{\eta \in \mathbf{v_k}} |\eta|$, respectively. Then action $\mathbf{v'_k} = s(i_k)$ splits the prior I-state equally into $i_k$ parts. Both the smallest and largest size of the result are equal to the average size:

$$\min_{\eta \in \mathbf{v'_k}} |\eta| = \max_{\eta \in \mathbf{v'_k}} |\eta| = \frac{|\eta_k| + \delta}{i_k}.$$

Since PPC and TTC bounds must be satisfied for all motions of the panda, the I-states arising from $\mathbf{v'_k}$, which are intermediate sized with $\min_{\eta \in \mathbf{v_k}} |\eta| \leq \frac{|\eta_k|+\delta}{i_k} \leq \max_{\eta \in \mathbf{v_k}} |\eta|$, cannot introduce a violation where none existed in $\pi$. ∎

Hence, instead of considering all possible choices for set-points $u_1 < u_2 < \cdots < u_c$, we only need to consider those in $\bigcup_{1 \leq i \leq c+1} s(i+1)$.

The actions can be categorized into two types depending on the number of parts the prior I-state is divided into. As shown in Figure 3, there are actions that increase the I-state size, and those that decrease its size. The distinguishing feature, which we denote with $a$, is the maximum number of $r_t$'s contained within the distance the panda can move in a single time-step, namely $a = \left\lceil \frac{\delta}{r_t} \right\rceil$. From this definition it follows that $\delta \in (ar_t - r_t, ar_t]$. Since $|\eta_k| \in [r_p, r_t]$, if $\delta \in [ar_p, ar_t]$, then action $s(a)$ will always guarantee that the I-state's size is admissible: $|\eta_{k+1}| = \frac{|\eta_k|+\delta}{a} \in [r_p, r_t]$. For $\delta \in (ar_t - r_t, ar_p)$, the size of the I-state that results from action $s(a)$ is $|\eta_{k+1}| = \frac{|\eta_k|+\delta}{a} > |\eta_k|$. The size of the resulting I-state under action $s(a+1)$ is $|\eta_{k+1}| = \frac{|\eta_k|+\delta}{a+1} < |\eta_k|$. Furthermore the size of the resulting I-state decreases with the parameter of the $s(\cdot)$ action. When the number partitions is less than or equal to $a$, the I-state size will increase; when the number of partitions exceeds $a$, the size of I-state will decrease.

Next, we reduce the action space into basis actions $s(a)$ and $s(a+1)$, which we find it useful to denote as $\oplus$ and $\ominus$, respectively.

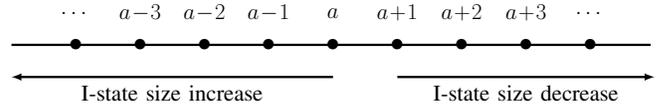

Fig. 3: The change of the resulting I-state's size when splitting into $i$ equal parts with $\delta \in (ar_t - r_t, ar_p)$.

**Lemma 2.** *For any 1-dim. problem $P_1 = (\eta_0, r_p, r_t, \delta, c)$, there exists a $\mathbf{PP}$ and $\mathbf{TT}$ strategy consisting solely of actions of $s(i)$ if and only if there exists a $\mathbf{PP}$ and $\mathbf{TT}$ strategy which uses actions $\oplus$ and $\ominus$ only.*

*Proof:* $\Leftarrow$: Holds as $\oplus = s(a)$ and $\ominus = s(a+1)$.
$\Rightarrow$: For any $\mathbf{PP}$ and $\mathbf{TT}$ strategy $\pi$ consisting of $s(i)$ actions, we construct $\pi'$ as follows. For each action $\mathbf{v_k}$ under strategy $\pi$, suppose $\mathbf{v_k}$ splits the prior I-state into $i_k$ parts. Construct the new action $\mathbf{v'_k}$ for $\pi'$, so that

$$\mathbf{v'_k} = \begin{cases} \oplus & \text{if } i_k \leq a, \\ \ominus & \text{if } i_k \geq a+1. \end{cases}$$

For the first case, where $i_k \leq a$, we have $r_p \leq |\eta_k| \leq \max_{\eta \in \mathbf{v'_k}} |\eta| \leq \max_{\eta \in \mathbf{v_k}} |\eta| \leq r_t$ And, thus one has that $\min_{\eta \in \mathbf{v'_k}} |\eta| = \max_{\eta \in \mathbf{v'_k}} |\eta| \in [r_p, r_t]$. For the second case, where $i_k \geq a+1$, this condition also holds similarly. The resulting I-state at each time-step under $\pi'$ satisfies both bounds, and $\pi'$ is a $\mathbf{PP}$ and $\mathbf{TT}$ strategy. ∎

**Theorem 1.** *For any tracking problem $P_1 = (\eta_0, r_p, r_t, \delta, c)$, $\mathbf{PP}(P_1) \wedge \mathbf{TT}(P_1)$ if and only if there exists a $\mathbf{PP}$ and $\mathbf{TT}$ strategy that consisting only of actions $\oplus$ and $\ominus$.*

*Proof:* Combine the results from Lemma 1 and 2. ∎

*C. Main results*

In this section, we follow the roadmap in Figure 2.

**Lemma 3.** *Let $P_1 = (\eta_0, r_p, r_t, \delta, c)$ be any 1-dim. panda tracking problem, if $\delta \in [ar_p, ar_t]$, where $a \in \mathbb{Z}^+$, $a \leq c$, then $\mathbf{PP}(P_1) \wedge \mathbf{TT}(P_1)$.*

*Proof:* A $\mathbf{PP}$ and $\mathbf{TT}$ strategy is given. For any $|\eta_k| \in [r_p, r_t]$ the prior I-state has size $|\eta^-_{k+1}| = |\eta_k| + \delta$. Since $\delta \in [ar_p, ar_t]$, where $a \leq c$, $|\eta^-_{k+1}| \in [ar_p + r_p, ar_t + r_t]$. By taking action $\ominus$, which is possible since $a \leq c$, we get $|\eta_{k+1}| = \frac{1}{a+1}|\eta^-_{k+1}| \in [r_p, r_t]$. That is, if $|\eta_0| \in [r_p, r_t]$ and we take action $\ominus$, then $\forall k, |\eta_k| \in [r_p, r_t]$. Therefore, there exists a strategy (always take action $\ominus$) for $P_1$, so that the privacy-preserving tracking conditions *PPC* and *TTC* are always both satisfied when $\eta_0 \in [r_p, r_t]$. ∎

**Lemma 4.** *For any 1-dim. panda tracking problem $P_1 = (\eta_0, r_p, r_t, \delta, c)$, if $\delta \in (cr_t, \infty)$, then $\neg \mathbf{TT}(P_1)$.*

*Proof:* The tracking stipulation is proved to be violated eventually. Given the constraint of $c$ sensing parameters, the prior I-state $|\eta^-_{k+1}| = |\eta_k| + \delta$ can be divided into at most $c+1$ parts. Among these $c+1$ posterior I-states, if $c$ of them reach the maximum size $r_t$, the size of the remaining I-state is $|\eta_k| + \delta - cr_t$. If none of the resulting I-states violate the tracking bound, then the size of the smallest resulting I-state is $|\eta_k| + \delta - cr_t$. Since $\delta > cr_t$, the size of the smallest resulting I-state must increase by some positive constant $\delta - cr_t$. After $\left\lceil \frac{r_t - r_p}{\delta - cr_t} \right\rceil$ stages, the I-state will exceed $r_t$. So it is impossible to ensure that the tracking bound will not eventually be violated. ∎

**Lemma 5.** *For 1-dim. panda tracking problem $P_1 = (\eta_0, r_p, r_t, \delta, c)$, if $\delta \in (ar_t - r_t, ar_p)$, where $a \in \mathbb{Z}^+$, $a \leq c$ and $ar_t \geq ar_p + r_p$, then $\mathbf{PP}(P_1) \wedge \mathbf{TT}(P_1)$.*

*Proof:* A **PP** and **TT** strategy is given in this proof. Since $\delta \in (ar_t - r_t, ar_p)$ and $ar_t > ar_p + r_p$, $|\eta_{k+1}^-| = |\eta_k| + \delta \in (ar_p, ar_t + r_t) \subset L_1 \cup L_2$, where $L_1 = [ar_p, ar_t]$ and $L_2 = [ar_p + r_p, ar_t + r_t]$. If action $\oplus$ is performed when $|\eta_{k+1}^-| \in L_1$ and $s(a+1)$ is performed when $|\eta_{k+1}^-| \in L_2$, the resulting I-state satisfies $|\eta_{k+1}| \in [r_p, r_t]$. Hence, there is a strategy consisting of $\oplus$ and $\ominus$, for the problem $P_1$ such that the *PPC* and *TTC* are always both satisfied when $\eta_0 \in [r_p, r_t]$. ∎

**Lemma 6.** *For any 1-dim. panda tracking problem $P_1 = (\eta_0, r_p, r_t, \delta, c)$, if $\delta \in (ar_t - r_t, ar_p)$, where $a \in \mathbb{Z}^+$, $a \leq c$, $ar_t < ar_p + r_p$, then $\neg \mathbf{PP}(P_1) \vee \neg \mathbf{TT}(P_1)$ when either: (i) $\delta > ar_t - r_p$ or (ii) $\delta < (a+1)r_p - r_t$.*

*Proof:* In case (i), $r_p > ar_t - \delta$, so the size of the prior I-state satisfies $|\eta_{k+1}^-| = |\eta_k| + \delta > r_p + \delta > ar_t$. That is, if we take action $\oplus$ and divide $\eta_{k+1}^-$ into $a$ parts, the largest posterior I-state $\eta_{k+1}$ will violate the tracking stipulation at the next time-step. If we take action $\ominus$ and divide $\eta_{k+1}^-$ into $a+1$ parts, it can be shown that the smallest I-state will eventually violate the privacy stipulation. Under the action $\ominus$, the smallest size of the resulting posterior I-state $|\eta_{k+1}|_{\text{smallest}} = \min_{\eta \in \mathbf{v_k}} |\eta|$ is no greater than the average size $\frac{|\eta_k| + \delta}{a+1}$. The amount of decrease is $\Delta_- = |\eta_k| - |\eta_{k+1}|_{\text{smallest}} \geq |\eta_k| - \frac{|\eta_k| + \delta}{a+1} \geq \frac{ar_p - \delta}{a+1} > 0$. Hence, eventually after $\left\lceil \frac{(a+1)(r_t - r_p)}{ar_p - \delta} \right\rceil$ steps, the smallest I-state will violate the privacy stipulation and put the panda in danger.

The same conclusion is reached for the case of (ii), when $r_t < (a+1)r_p - \delta$, along similar lines. ∎

Let the domain of $\delta$ described by condition (i) in Lemma 6 be $K_1 = (ar_t - r_t, ar_p) \cap (ar_t - r_p, +\infty)$, and that of condition (ii) be $K_2 = (ar_t - r_t, ar_p) \cap (-\infty, (a+1)r_p - r_t)$. If $(a+1)r_t < (a+2)r_p$, then $ar_t - r_p < (a+1)r_p - r_t$. We have $K_1 \cup K_2 = (ar_t - r_t, ar_p)$, so condition (i) and (ii) together describe problems making up all the gaps in Figure 2, save for the last and penultimate one. The previous lemma shows that there are no privacy-preserving strategies for those gaps.

Notice that Lemma 3 and 5 describe under-constrained problems where there exists a straightforward privacy-preserving tracking strategy. Lemma 4 and 6 prove that there are no privacy-preserving tracking strategies for a set of problems which we might consider to be over-constrained. Next, we give a lemma that describes boundary problems separating under-constrained from over-constrained problems. (Definition 5 in Section V formalizes these boundary problems, for here it is sufficient to recognize that these are elements from the penultimate gap in Figure 2.)

**Lemma 7.** *For any 1-dim. panda tracking problem $P_1 = (\eta_0, r_p, r_t, \delta, c)$, if $a \in \mathbb{Z}^+$, $a \leq c$ and $r_p \leq ar_t - \delta < (a+1)r_p - \delta \leq r_t$, then there is no privacy-preserving strategy for every initial I-state.*

*Proof:* For $|\eta_0| \in (ar_t - \delta, (a+1)r_p - \delta)$, if we take action $\oplus$ and equally divide the prior I-state into $a$ parts, the size of the largest posterior I-state at the next time-step ($t = 1$) is $|\eta_1| = \frac{|\eta_1^-| + \delta}{a} > \frac{ar_t - \delta + \delta}{a} = r_t$, which will violate the tracking stipulation. If we take action $\ominus$ and equally divide the prior I-state into $a + 1$ parts, the size of posterior I-state at the next step ($t = 1$) is $|\eta_1| = \frac{|\eta_1^-| + \delta}{a+1} < \frac{(a+1)r_p - \delta}{a+1} = r_p$, which will violate the privacy stipulation. Hence, there are no strategies, which are both **PP** and **TT**, for $|\eta_0| \in (ar_t - \delta, (a+1)r_p - \delta)$. ∎

The preceding proof shows that, in the case of these problems, there are no strategies which are both **PP** and **TT** for all initial I-states. It achieves this by showing that there are particular initial I-states for which either the PPC or the TTC must be violated. But the problems in Lemma 7 are not merely over-constrained instances—something rather more complex is going on in these boundary problems. To illustrate this fact, next, we show that there exist boundary problems that have a **PP** and **TT** strategy but only for certain initial I-states.

**Example 1.** *Consider $Q_1 = (\eta_0, r_p = 76, r_t = 101.3, \delta = 227, c = 4)$. It satisfies the constraints in Lemma 7 (viz., $r_p \leq ar_t - \delta < (a+1)r_p - \delta \leq r_t$ and $a \leq c$, since $a = 3$) and also has privacy-preserving strategies described by*

$$\pi_{Q_1} = (\oplus(\ominus)^3)^*$$

*for $|\eta_0| \in [76, 76.9]$. The asterisk in the expression for $\pi_{Q_1}$ above should be interpreted as a Kleene star. By removing the appropriate prefix of actions, $\pi_{Q_1}$ can also be extended to work for any $|\eta_0| \in [76, 76.9] \cup [77, 80.6] \cup [81, 95.4] \cup [97, 101.3]$.*

This problem instance and the behavior of the strategy is illustrated in Figure 4.

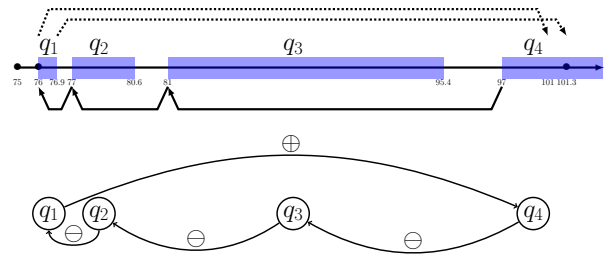

Fig. 4: The problem instance given in Example 1, where there exists a **PP** and **TT** tracking strategy for $|\eta_0| \in q_1 \cup q_2 \cup q_3 \cup q_4$.

**Example 2.** *We can adapt $Q_1$ from Example 1 by considering a slightly more lethargic panda, which has $\delta = 223$. This*

gives a new problem instance $Q'_1 = (\eta_0, r_p = 76, r_t = 101.3, \delta = 223, c = 4)$, which has more complex privacy-preserving strategies, again written in regular expression-like form

$$\pi_{Q'_1} = (\oplus \ominus \oplus (\ominus)^2)^*$$

for $|\eta_0| \in [76, 79.21]$. As with the case above, after dropping an appropriate prefix of actions, $\pi_{Q'_1}$ will work on initial states: $|\eta_0| \in [76, 79.21] \cup [80, 80.9] \cup [81, 93.86] \cup [96.66, 100.6] \cup [101, 101.3]$.

And this problem instance and how the strategy $\pi_{Q'_1}$ relates is illustrated in Figure 5.

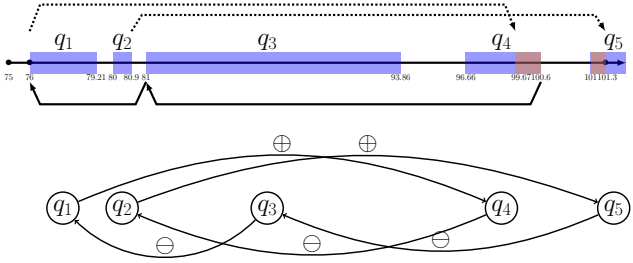

Fig. 5: The problem instance given in Example 2, where there exists a **PP** and **TT** tracking strategy for $|\eta_0| \in q_1 \cup q_2 \cup q_3 \cup q_4 \cup q_5$.

Both Figures 4 and 5 are instances of problems whose feasibility depends on the size of the initial I-state, $|\eta_0|$. They illustrate that, unlike the under-constrained and over-constrained problems, any conclusions reached about such instances demand a deeper analysis—the details of that analysis are deferred till Section V. First, we collect the results presented thus far, clarifying their interplay and giving some interpretation.

### D. An aggregate visualization of results that are independent of initial I-state

To have a clearer sense of how these pieces fit together, we have found it helpful to plot the space of problem parameters and examine how the preceding theorems relate visually. Figure 6 contains subfigures for increasingly powerful robots (in terms of sensing) with $c = 1, 2, 3, 4$. The white regions represent the trivial $r_p > r_t$ instances; otherwise the whole strategy space is categorized into the following subregions: under-constrained space (colored green), over-constrained space (colored gray), and boundary space (colored pink). When summarized in this way, the results permit examination of how sensor power affects the existence of solutions.

Preserving the privacy of a target certainly makes some unusual demands of a sensor. O'Kane's quadrant sensor has preimages for each output class that are infinite subsets of the plane, making it possible for his robot to increase its uncertainty if it must do so. But it remains far from clear how one might tackle the same problem with a different sensor. The privacy requirement makes it difficult to reason about the relationship between two similar sensors. For example, an octant sensor appears to be at least as useful as the quadrant sensor, but it makes preserving privacy rather trickier. Since octants meet at the origin at 45°, it is difficult to position the robot so that it does not discover too much. One advantage of the one-dimensional model is that the parameter $c$ allows for a natural modification of sensor capabilities. This leads to three closely related results, each of which helps clarify how certain limitations persist even when $c$ is increased.

**Theorem 2.A.** *(More sensing won't grant omnipotence)* The one-dimensional robot is not always able to achieve privacy-preserving tracking, regardless of its sensing power.

*Proof:* The negative results in the over-constrained problems described by Lemmas 4 and 6 show that there are circumstances where it is impossible to find a tracking strategy satisfying both *PPC* and *TTC*. Though these regions depend on $c$, no finite value of $c$ causes these regions to be empty. ∎

Turning to the boundary cases that serve as the transition from over-constrained problems to under-constrained ones, one might think that these boundary regions will shift or reduce when the sensor gets more powerful to handle the constraints. But this explanation is actually erroneous. Observe that the boundary region is more complicated than other regions within the strategy space: in Figure 6, the green region is contiguous, whereas the regions marked pink are not. The specific boundary region for Lemma 7 under condition $a = 1$, visible clearly as chisel shape in Figures 6a and 6b, is invariant with respect to $c$, so remains as the boundary in all circumstances (though outside the visible region in Figures 6c and 6d, it is present). As $c$ increases, what happens is that the former boundary regions still serve as the boundary, and the regions previously marked as over-constrained are claimed as under-constrained, or become the boundary regions. It is evident that the boundary regions do not shift with more powerful sensors. The following expresses the fact that additional sensing power fails to reduce the number of the boundary regions.

**Theorem 2.B.** *(Boundary region invariance)* The number of boundary regions does not decrease by using more powerful sensors.

*Proof:* We focus on the boundary areas within the square between $(0, 0)$ and $(2, 2)$ as the parts outside this (orange) square do not change as $c$ increases. According to Lemma 7, the boundary regions for any specific $a$ are bounded by the following linear inequalities:

$$r_t < \frac{2}{a-1}, \quad (1)$$

$$r_p > \frac{2}{a}, \quad (2)$$

$$r_t \geq \frac{r_p}{a} + \frac{2}{a}, \quad (3)$$

$$r_t \geq (a+1)r_p - 2, \quad (4)$$

$$r_t < \frac{(a+1)}{a}r_p. \quad (5)$$

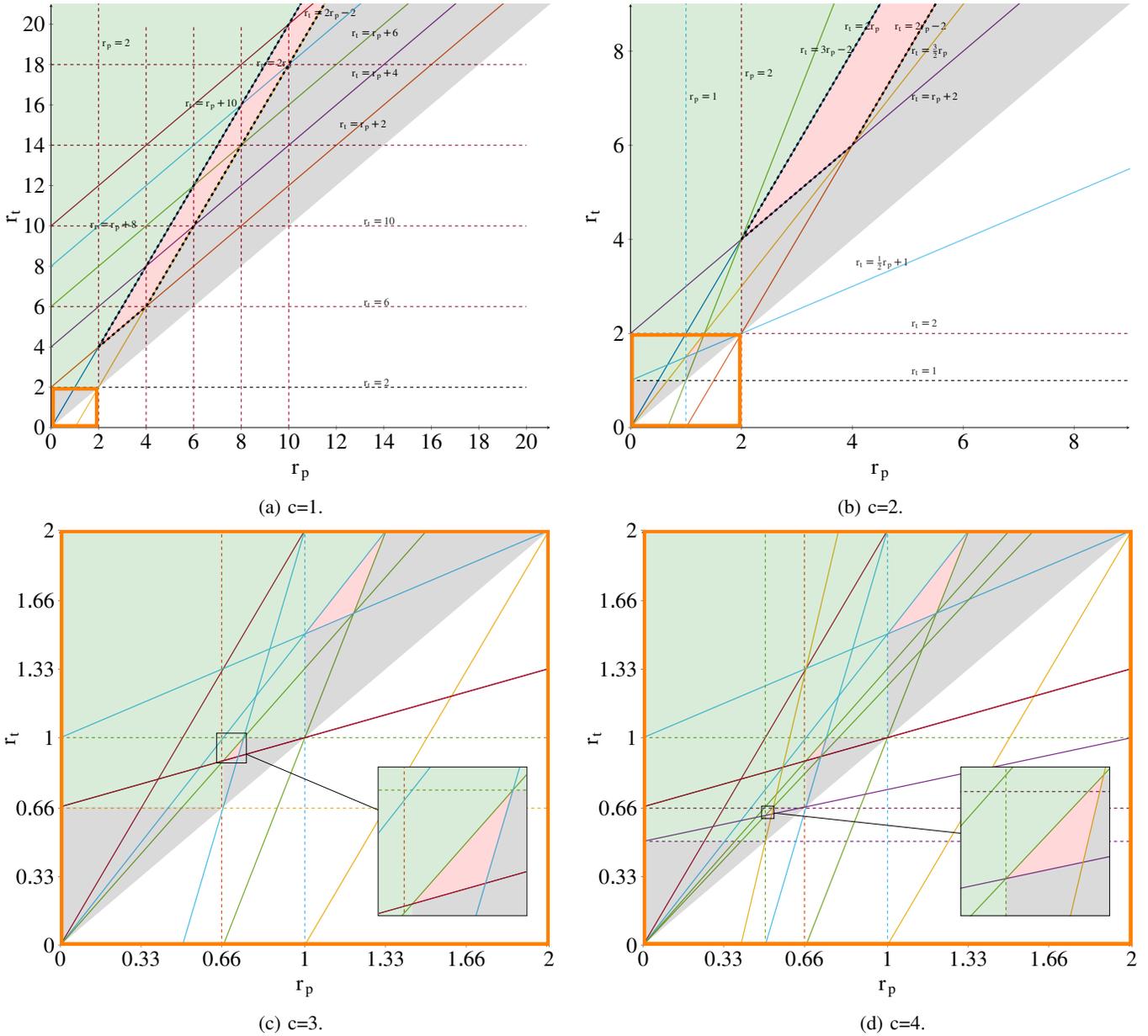

Fig. 6: The problem parameter space and the existence of strategies for robots with differing $c$. The green region depicts under-constrained problems, where a suitable strategy exists no matter the initial I-state. The gray region represents the conditions that are over-constrained. The pink region depicts conditions that serve as the boundary between over-constrained and under-constrained problems. The white region represents trivially infeasible problems. The orange rectangles emphasize the different levels of magnification and highlight conditions where the differences in sensing power come into play. Both $r_p$ and $r_t$ are expressed in units of $\frac{\delta}{2}$.

Combining (1)–(3) gives both the bound for $r_t$ as $r_t \in \left[\frac{2(a+1)}{a^2}, \frac{2}{a-1}\right)$, and the bound for $r_p$ as $r_p \in \left[\frac{2}{a}, \frac{2a}{a^2-1}\right]$. The boundary region, thus, is bounded.

Next, we show that (1) and (2) are dominated by (3)–(5). According to (4) and (5), we have $r_p < \frac{2a}{a^2-1}$. Applying this result to (5) produces (1). Similarly, combining (3) and (5) together yields (2). Hence, the boundary regions are fully determined by inequalities (3)–(5). (Figure 7 provides a visual example.)

To form a bounded region with three linear inequalities, the boundary region has to be a triangle. The three points of the triangle can be obtained by intersecting pairs of (3)–(5): $\left(\frac{2}{a}, \frac{2a+2}{a^2}\right)$, $\left(\frac{2a}{a^2-1}, \frac{2}{a-1}\right)$, $\left(\frac{2(a+1)}{a^2+a-1}, \frac{2(a+1)^2}{a^2+a-1} - 2\right)$. Since $a \in \{2, 3, \cdots, c\}$, the triangle region will not be empty. Let $\Delta(a)$ denote the triangle with parameter $a$. Then the smallest $y$ coordinate for $\Delta(a)$ is $\min Y(\Delta(a)) = \frac{2a+2}{a^2}$. And the largest $y$ coordinate for $\Delta(a)$ is $\max Y(\Delta(a)) = \frac{2}{a-1}$. For adjacent triangles $\Delta(a)$ and $\Delta(a+1)$, we have $\min Y(\Delta(a)) > \max Y(\Delta(a+1))$. Hence, the triangles for different values of $a$ do not overlap. ∎

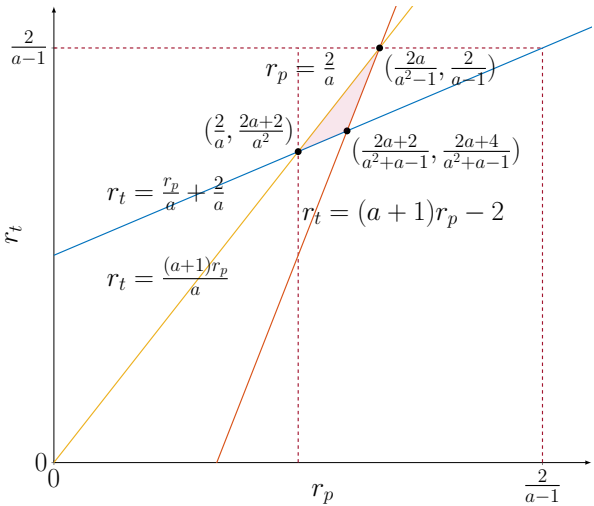

Fig. 7: Relationships between the linear inequalities in Lemma 7.

The preceding discussion showed that the number of boundary regions increases with $c$ and that, along with the chisel shaped region, there are $c - 1$ triangles of decreasing size. This motivates our introduction of a quantitative measure of the robot's power as a function of $c$.

**Definition 4.** A measure of tracking power, $p(c)$, for the robot with $c$ sensing parameters should satisfy the following two properties: $p(c) > 0$, $\forall c \in \mathbb{Z}^+$ (*positivity*), and $p(a) > p(b)$, if $a, b \in \mathbb{Z}^+$ and $a > b$ (*monotonicity*).

The plots in Figure 6 suggest that one way to quantify change in these regions is to measure changes in the various areas of the parameter space as $c$ increases. As a specific measure of power in the one-dimensional setting, we might consider the proportion of cases (in the $r_p$ vs. $r_t$ plane) that are under-constrained (green) and boundary regions (pink). Though the green and pink areas are unbounded in the full plane, the only changes that occur as $c$ increases are in the square between $(0, 0)$ and $(2, 2)$. Thus, we take $p(c)$ to equal to the total volume of green and pink regions filling within the region $0 \leq r_p \leq r_t$ and $0 \leq r_t \leq 2$. This area satisfies the properties in Definition 4 and is indicative of the power of the robot as, intuitively, it can be interpreted as an upper-bound of the solvable cases.

**Corollary 1.** *(Asymptotic tracking power)* The power $p(c)$ of a robot with $c$ sensing parameters to achieve privacy-preserving tracking in the 1-dim. problem is bounded and $\lim_{c \to \infty} p(c) = \ell$, with $1.5 < \ell < 1.6$.

*Proof:* Inequalities (3)–(5) in the proof of Theorem 2.B give $c - 1$ triangles, one for each $a \in \{2, 3, \cdots, c\}$. An analytic expression gives the area of each of these triangles and the series describing the cumulative pink volume $p_{\text{pink}}(c)$ within $0 \leq r_p \leq r_t$ and $0 \leq r_t \leq 2$ can be shown (by the comparison test) to converge as $c \to \infty$. Similarly, the cumulative green volume $p_{\text{green}}(c)$ within $0 \leq r_p \leq r_t$ and $0 \leq r_t \leq 2$ converges. Numerical evaluation gives the value of the limit $\approx 1.54\bar{5}$. ∎

*E. The greatest tracking precision for a given problem instance*

A practical question what one might ask is: for a given set of panda tracking parameters (the panda's motion $\delta$, privacy bound $r_p$, sensor capability $c$), what is the highest precision (i.e., tightest tracking bound), for which the problem is solvable? Formally this requires us to find the smallest $r_t$, which we denote $r_t^\star$, such that an **PP** and **TT** tracking strategy exists $\forall \eta_0, P_1 = (\eta_0, r_p, r_t^\star, \delta, c)$. For small values of $c$, the lower envelope of the green region in Figure 6 shows the values of $r_t^\star$ graphically. By tracing the positions of the under-constrained problems, over-constrained ones, and boundary regions, an answer to this question can be obtained more generally with the following formula (where we have let $c^\star = \lceil \frac{\delta}{r_p} \rceil$):

$$r_t^\star = \begin{cases} 2r_p, & r_p \geq \delta, \\ \delta, & c = 1, r_p < \delta, \\ \frac{\delta}{c}, & 1 < c < c^\star, r_p < \delta, \\ \frac{(c^\star + 1)r_p}{c^\star}, & c \geq c^\star, r_p \in [\frac{\delta}{c^\star}, \frac{\delta c^\star}{c^{\star 2} - 1}], r_p < \delta, \\ \frac{\delta}{c^\star - 1}, & c \geq c^\star, r_p \in (\frac{\delta c^\star}{c^{\star 2} - 1}, \frac{\delta}{c^\star - 1}), r_p < \delta. \end{cases}$$

The conditions on the right-hand side are exhaustive.

## V. BOUNDARY PROBLEMS SHOW A DEPENDENCE ON THE ROBOT'S INITIAL I-STATE

The results of the preceding section provide an analysis for the green and gray areas in Figure 6, which correspond to the teeth, gaps, filling, and final interval in the roadmap diagram (Figure 2). Only the penultimate interval, resulting in pink regions in Figure 6, is left. In this section we analyze these last remaining problem instances, that is the boundary problems, and they are shown to have a different structure from the others (which is why we have opted to treat them separately). We show that there are no **PP** and **TT** tracking strategies for all initial I-states in these instances but, as will be uncovered, there do exist some boundary problems that have **PP** and **TT** tracking strategies for some initial I-states.

First, we describe the common characteristics of boundary problems, introducing the notion of an "impossibility zone", thereafter we give a detailed treatment of the each of the classes of problems that arise (Lemmas 9–12), The last subsection dealing with boundary problems provides some broader interpretation of the technical results.

### A. The boundary problems in detail

The boundary problem described is defined as follows:

**Definition 5.** The 1-dim. panda tracking problem $P_1 = (\eta_0, r_p, r_t, \delta, c)$ is a *boundary problem*, if $\delta \in (ar_t - r_t, ar_p)$, $r_p \leq ar_t - \delta < (a+1)r_p - \delta \leq r_t$, and $a \in \{1, 2, \ldots, c\}$.

In boundary problem $P_1$, if the size of the I-state is $ar_t - \delta$, then it will transit to the I-state with size $r_t$ under action $\ominus$. The I-state with size $(a+1)r_p - \delta$ will transit to the I-state with size $r_p$. Between $ar_t - \delta$ and $(a+1)r_p - \delta$, there is a zone $I_0^\times = (ar_t - \delta, (a+1)r_p - \delta)$ shown in Figure 8, wherein all actions will violate the privacy or tracking bound in the next time-step. We call $I_0^\times$ the *impossibility zone*. To its left, $\oplus$ is the only action that can be taken, otherwise the privacy constraint will be violated in the next time-step. Similarly, $\ominus$ is the only action that can be taken to the right of the impossibility zone. The robot's actions are forced, and making any other choice means that the privacy or tracking constraints will be violated immediately afterward. Thus we may conclude that if there is a **PP** and **TT** strategy for some $\eta_0$, then the privacy-preserving tracking strategy is unique.

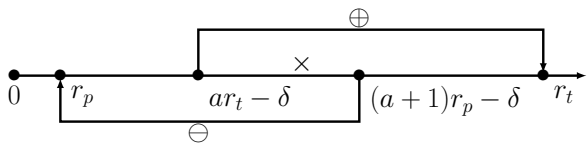

Fig. 8: The boundary problem instances have the property that there is a central zone from which no action can safely be taken thereafter.

Next, we will examine the I-state transitions under action $\oplus$ and $\ominus$, so as to understand the structure of the privacy-preserving tracking strategy in the boundary problem. Let the resulting size of I-state after taking action $\oplus$ and $\ominus$ be $f_+(x)$ and $f_-(x)$. For a perturbation $\Delta_x$, the two functions satisfy the following:

1) If $\Delta_x > 0$, then $f_+(x + \Delta_x) - f_+(x) = \frac{\Delta_x}{a} > 0$.
2) If $\Delta_x > 0$, then $f_-(x + \Delta_x) - f_-(x) = \frac{\Delta_x}{a+1} > 0$.

These properties help in understanding the transition of I-states under action $\oplus$ and $\ominus$. Both $f_+$ and $f_-$ are monotonically increasing functions, preserving the order of their inputs. If the same action is performed at the both endpoints of some interval $[x, x+\Delta_x]$, then the interval transits to $[f_+(x), f_+(x+\Delta_x)]$ under action $\oplus$, and $[f_-(x), f_-(x + \Delta_x)]$ under action $\ominus$. Extending the notation naturally, we will write the interval $[f_+(x), f_+(x + \Delta_x)]$ as $f_+([x, x + \Delta_x])$. The size of the new interval will decrease to $\frac{1}{a}$ of the original interval under action $\oplus$, and $\frac{1}{a+1}$ of the original one under action $\ominus$. That is, $|f_+([x, x + \Delta_x])| = \frac{\Delta_x}{a}$ and $|f_-([x, x + \Delta_x])| = \frac{\Delta_x}{a+1}$.

Following these transition properties, we are able to divide $[r_p, r_t]$ into subintervals. Let $p$ be the maximum number of sequential $\oplus$'s that can be performed before violating the tracking constraint $r_t$ for all I-states, and $m$ be the maximum number of sequential $\ominus$'s that before violating the privacy constraint $r_p$ for all I-states. Then, as shown in Figure 9, the interval $[r_p, r_t]$ can be divided into three parts: $\bigcup_{1 \leq j \leq p} I_j^+$, the impossibility zone $I_0^\times$, and $\bigcup_{1 \leq j \leq m} I_j^-$. For each interval $I_j^+ = (l_j^+, r_j^+]$, we have $l_{j+1}^+ = r_j^+$ and $l_j^+ = f_+^{-1}(r_j^+)$, where $f_+^{-1}(x)$ denotes the size of the I-state that transits to a new one of size $x$ after taking action $\oplus$. The "+" in $I_j^+$ means that only action $\oplus$ can be taken, and $j$ is the number of sequential $\oplus$'s that can be taken before violating the tracking constraints. Similarly, for any interval $I_j^- = [l_j^-, r_j^-]$, we have $r_{j+1}^- = l_j^-$ and $r_j^- = f_-^{-1}(l_j^-)$. The "-" in $I_j^-$ means that only action $\ominus$ can be taken, and there are at most $j$ sequential $\ominus$'s to be taken before violating the privacy constraints.

Following from the I-state transition properties and as Lemma 8 states, the maximum number of $\oplus$'s and $\ominus$'s are constrained.

**Lemma 8.** *In the boundary problem $P_1$, either $m = 1$ or $p = 1$.*

*Proof:* The proof is by contradiction by assuming that both $m > 1$ and $p > 1$. Since interval $I_1^+$ transits to a new one containing $I_1^-$ under $\oplus$, we can conclude that $|I_1^+| > |f_+(I_1^+)| > |I_1^-|$. But interval $I_1^-$ transits to a new one containing $I_1^+$ after action $\ominus$. Hence, $|I_1^-| > |f_-(I_1^-)| > |I_1^+|$, which contradicts the prior assertion. Therefore the assumption is incorrect and either $m = 1$ or $p = 1$. ∎

There are two cases, shown in Figure 10, depending on whether $p = 1$ or $m = 1$. We see that the unique privacy-preserving tracking strategy consists of $\oplus$ and $\ominus$ in the following regular expression form $\oplus^*(\ominus(\oplus)^{p|p-1})^*$ or $\ominus^*(\oplus(\ominus)^{m|m-1})^*$, where the prefixes on the action outside the parentheses have length no more than the same one inside.

In boundary problems, the impossibility zone, $I_0^\times$, contains the I-states that will violate the privacy or tracking bound in one time-step for any action. The next step is to define $I_j^\times$, where $j \in \mathbb{N} \cup \{0\}$, such that $I_j^\times$ represents the set of I-states that will violate the privacy or tracking bound in $j + 1$ steps. Then $I_j^\times$ is defined (recursively) as:

$$I_j^\times = \begin{cases} f_+^{-1}(I_{j-1}^\times) \cap [r_p, r_t] & \text{if } f_-^{-1}(I_{j-1}^\times) \cap [r_p, r_t] = \varnothing, \\ f_-^{-1}(I_{j-1}^\times) \cap [r_p, r_t] & \text{if } f_+^{-1}(I_{j-1}^\times) \cap [r_p, r_t] = \varnothing. \end{cases}$$

The two conditions are mutually exclusive because actions are forced for boundary problems: we have either $f_-^{-1}(i_{j-1}^\times) \cap [r_p, r_t] = \varnothing$ or $f_+^{-1}(i_{j-1}^\times) \cap [r_p, r_t] = \varnothing$, since at least one action is illegal, taking $I_{j-1}^\times$ out of $[r_p, r_t]$. But the two conditions need not be jointly exhaustive and, if neither condition holds for some $j$, then no such $I_j^\times$ is defined.

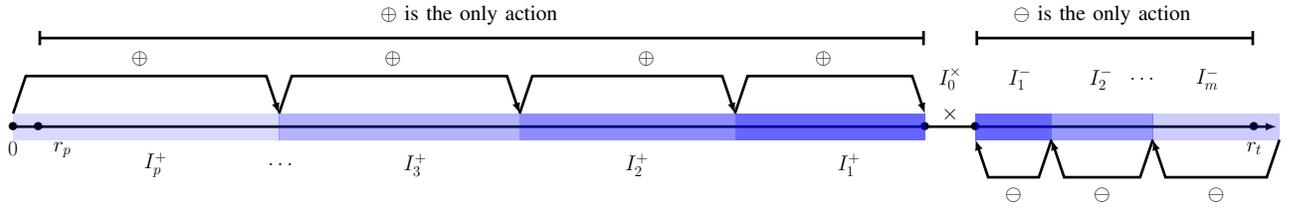

Fig. 9: $m$ $\oplus$'s and $p$ $\ominus$'s can be taken in $[r_p, ar_t - \delta]$ and $[(a+1)r_p - \delta, r_t]$.

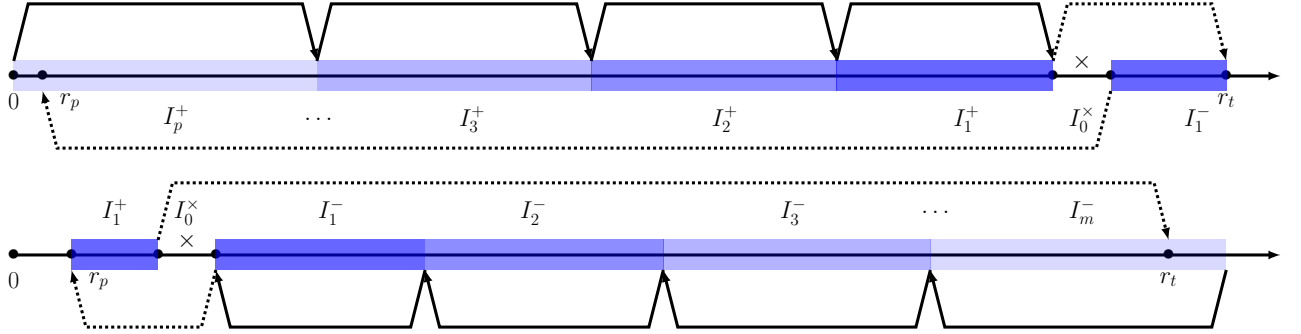

Fig. 10: All boundary problem instances have either a single $\ominus$ action (top) or single $\oplus$ action (bottom).

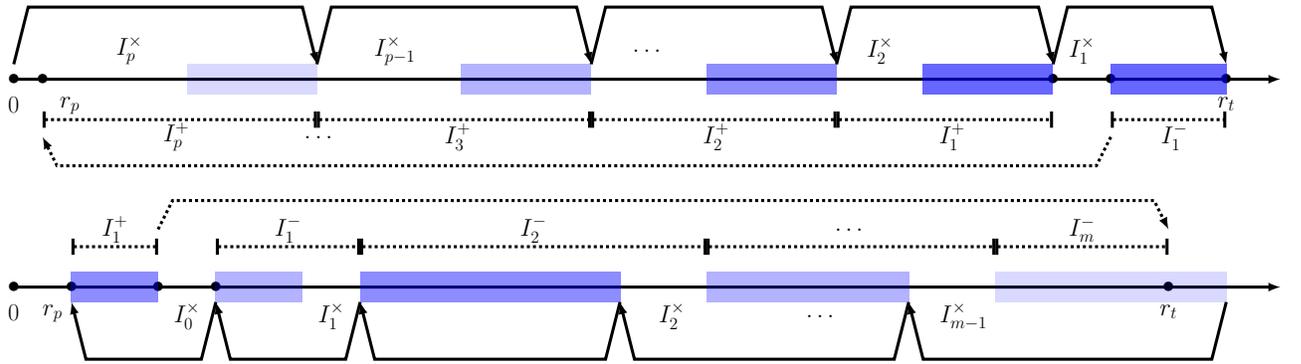

Fig. 11: Two cases of the boundary problem after the propagating the impossibility zone with $\oplus^*$ or $\ominus^*$. The top instance fits the description in the text.

For every $I_j^\times$ we are justified in calling it an interval, for it is a single interval, and $\exists k \in \{1, 2, \dots\}$, such that $I_j^\times \subseteq I_k^+$ or $I_k^-$. Both of these properties are easily shown to hold inductively (though the $m = 1$ and $p = 1$ cases, see Figure 10, each demand a slightly different inductive basis).

To find all the impossibility zones, we need to trace through the I-state transitions until no new impossibility zones are produced, i.e., until neither condition holds. Notice that the definition is in terms of the inverse maps of $f_+$ and $f_-$ so, instead of following the I-state transition forwards, we must proceed backwards to identify the new impossibility zone $I_{j+1}^\times$ that transits to $I_j^\times$. We call this process of tracing inverses back-propagation. In the first case ($m = 1$), the propagation advances backward using the inverse of $\oplus^*(\ominus(\oplus)^{p|p-1})^*$, and for ($p = 1$), the second case, by using the inverse of $\ominus^*(\oplus(\ominus)^{m|m-1})^*$. Since the two cases are symmetric, from this point onwards we examine the first only, though all properties also hold for the second and can be derived in an analogous manner.

The back-propagation of impossibility zones is periodic and each period can be divided into two phases. The first phase is the back-propagation from $I_1^-$ to $I_p^+$ or $I_{p-1}^+$ under the inverse of action sequence $\oplus^*$. The second is the back-propagation from $I_p^+$ or $I_{p-1}^+$ to $I_1^-$ following the inverse of action $\ominus$. The first phase is straightforward, since the fraction of the impossibility zone in $I_j^+$ during the propagation is the same as that of $I_1^- \cup I_0^\times$ in the same period. A visual example of back-propagation in the first phase of the first period appears in Figure 11. Depending on the size of interval $[r_p, r_t]$, there are either $p$ or $p + 1$ impossibility zones. The second phase is important, since it determines the fraction in $I_1^-$ in the next period. To satisfy the constraint that $|f_-(I_1^-)| < |I_1^-| < |I_{p-1}^+|$, there are four possible transitions as shown in Figure 12. Each transition maps $f_-(I_1^-)$ into different parts of $I_p^+ \cup I_{p-1}^+$, which results in a different fraction in the next period. As one continues to follow the periodic transition there are two outcomes: (i) if the fraction of the impossibility zone in $I_1^-$ converges to a value less than 1, then we have found all the impossibility zones and the I-states that are not in the impossibility zone have a **PP** and **TT** strategy; (ii) if the impossibility zone fills $I_1^-$, then the impossibility zones taint the whole $[r_p, r_t]$ and no regions with privacy-preservable tracking strategies remain.

### B. The impossibility zone's propagation

To track of the fraction of the impossibility zone in $I_1^-$ at each period, we denote the fraction at period $t$ as $z_t$. In the first period, there is no impossibility zone in $I_1^-$ and $z_1 = 0$. In Lemmas 9–12, we will compute $z_t$ for each possible critical transition.

**Lemma 9.** *For the boundary problem $P_1 = (\eta_0, r_p, r_t, \delta, c)$, if $r_p \in I_p^\times$, then there are no **PP** and **TT** strategies for any $\eta_0 \in [r_p, r_t]$.*

*Proof:* The impossibility zone is proved to ultimately taint the entire interval $[r_p, r_t]$. If $r_p \in I_p^\times$, then $f_-^{-1}(I_1^-)$ is mapped into two parts: the impossibility part $B_1 = f_-^{-1}(I_1^-) \cap I_p^\times$ and the non-impossibility part $f_-^{-1}(I_1^-) \cap (I_p^+ \setminus B_1)$. Following this mapping, the fraction of the impossibility zone $I_{p+1}^\times$ in $I_1^-$ in the next period, $t = 2$, is $z_2 = \frac{(a+1)|B_1|}{|I_1^-|}$. In the first phase of period $t = 2$, the impossibility zone $I_{p+1}^\times$ will be back-propagated to $I_p^+$ as $I_{2p+1}^\times$, which taints part of the non-impossibility part of the previous period and gets mapped to $I_1^-$ in the next period. There is a relationship between the sizes: $|I_{2p+1}^\times| = a^p(a+1)|B_1|$. Let $B_2$ be the additional impossibility interval that will be back-propagated in the next period. But $B_1$ and $B_2$ are contiguous, since $B_2$ is the impossibility zone to the left of $I_p^+$. Then $z_3 = (a+1)\frac{|B_1|+|B_2|}{|I_1^-|}$, where $|B_2| = a^p(a+1)|B_1|$. Following this back-propagation pattern in general, at period $t$, the additional impossibility zone to be back-propagated to $I_1^-$ is $B_{t-1}$, where $|B_{t-1}| = a^p(a+1)|B_{t-2}|$ and, again, $B_{t-1}$ is contiguous with $B_{t-2}$. Therefore, after some (finite) time $\kappa$, $z_\kappa$ will reach 1, and the impossibility zone will have tainted the whole interval $[r_p, r_t]$. ∎

To sum up visually, Lemma 9 shows that there are no **PP** and **TT** strategies in the boundary problem described by Figure 12a.

In the following I-state transitions, there are now only $p - 1$ impossibility zones that can be found during the first phase of the first period. Let these impossibility zones be

$$L = \bigcup_{0 \leq j \leq p-1} I_j^\times.$$

**Lemma 10.** *For the boundary problem $P_1 = (\eta_0, r_p, r_t, \delta, c)$, if $r_p \in I_p^+$ and $f_-(r_t) \in I_p^+$, then $P_1$ is I-state dependent: there are **PP** and **TT** strategies for $|\eta_0| \in (I_p^+ \setminus L)$, while there are no such strategies for $|\eta_0| \in L$.*

*Proof:* The fraction of the impossibility zone within $I_1^-$ reaches a maximum fraction less than 1, and a **PP** and **TT** strategy for $|\eta_0| \in ([r_p, r_t] \setminus L)$ is given in this proof. If $r_p \in I_p^+$ and $f_-(r_t) \in I_p^+$, then $f_-(I_1^-)$ is mapped into the $[r_p, r_t] \setminus L$, and no impossibility will back-propagate to $I_1^-$ at period $t = 2$ or afterward. Hence, the fraction of the impossibility zone in $I_1^-$ remains zero and, therefore, there is no **PP** and **TT** strategy for $L$, but a privacy-preserving tracking strategy for $[r_p, r_t] \setminus L$ exists. The privacy-preserving tracking strategy for $[r_p, r_t] \setminus L$ is as follows: take action $\oplus$ in $[r_p, r_t] \setminus (L \cup I_1^-)$, and action $\ominus$ in $I_1^-$. The resulting I-state remains within $[r_p, r_t] \setminus L$ forever. ∎

Lemma 10 says that there are **PP** and **TT** strategies in the boundary problem described by Figure 12b.

**Lemma 11.** *For the boundary problem $P_1 = (\eta_0, r_p, r_t, \delta, c)$, there are no **PP** and **TT** strategies for any $\eta_0 \in [r_p, r_t]$ when $r_p \in (I_p^+ \setminus L)$ and $f_-(r_t) \in I_{p-1}^\times$.*

*Proof:* The proof is similar to Lemma 9. ∎

Lemma 11 says that there are no **PP** and **TT** strategies in the boundary problem described by Figure 12c.

For the fourth type of I-state transition, since $r_p \in I_p^+$ and $f_-(r_t) \in (I_{p-1}^+ \setminus I_{p-1}^\times)$, it follows that $f_-(I_1^-)$ is mapped to three parts: the first non-impossibility zone $f_-(I_1^-) \cap (I_p^+ \setminus L)$, impossibility zone $I_p^\times$, and the second non-impossibility zone $f_-(I_1^-) \cap (I_{p-1}^+ \setminus L)$. Following the forward I-state transition, part of first one non-impossibility zone will transit to part of the second one and vice versa. It turns out that the fraction of impossibility zones in $I_1^-$ stays the same, remaining less than 1, when the size of the first and the second non-impossibility zone are comparable in size. And there is, thus, a **PP** and **TT** strategy for some initial I-states in this condition. Otherwise, there is no privacy-preserving tracking strategy for all of $[r_p, r_t]$. This conclusion is reached by first defining $v = f_-(I_1^-) \cap I_p^+$ as the first non-impossibility part, and $w = f_-(I_1^-) \cap I_{p-1}^+$ as the second non-impossibility part. Details are in the following proof for Lemma 12.

**Lemma 12.** *For the boundary problem $P_1$ with $r_p \in I_p^+$ and $f_-(r_t) \in (I_{p-1}^+ \setminus I_{p-1}^\times)$:*

1) *If*

$$\frac{1}{a^p(a+1)} \leq \frac{|w|}{|v|} \leq a^{p-1}(a+1),$$

*then $P_1$ is initial I-state dependent: there is a **PP** and **TT** strategy for the initial I-states $\eta_0 \in [r_p, r_t] \setminus L$, while there are no **PP** and **TT** strategies for $I_j^\times \in L$.*

2) *If either $\frac{|w|}{|v|} > a^{p-1}(a+1)$ or $\frac{|w|}{|v|} < \frac{1}{a^p(a+1)}$, then there are no **PP** and **TT** strategies for $|\eta_0| \in [r_p, r_t]$.*

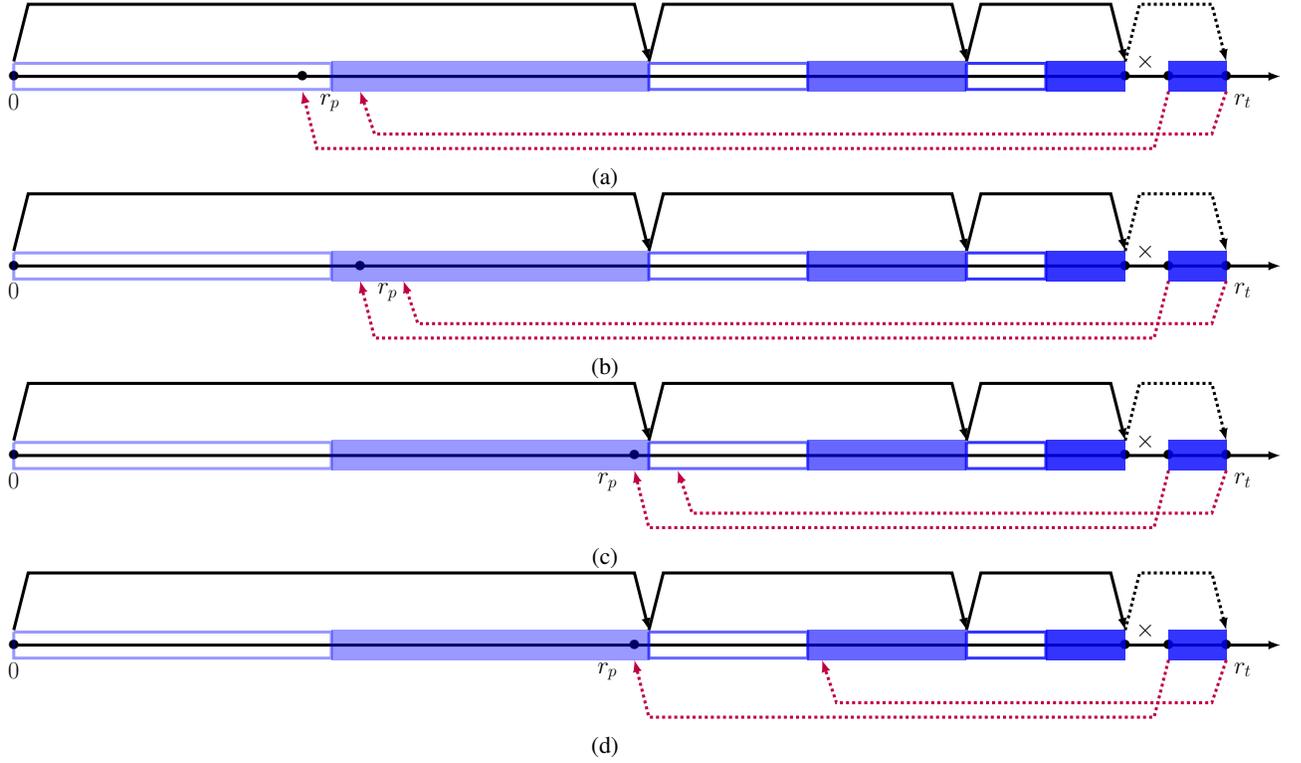

Fig. 12: Four possibilities of the I-state transition for $I_1^-$ in the first case of the boundary problem. These correspond to Lemmas 9–12.

*Proof:* First we prove that if $\frac{|w|}{|v|} > a^{p-1}(a+1)$, the impossibility zones propagate until they taint all of $[r_p, r_t]$. By tracking I-state transitions the impossibility zone in the middle, $I_{p-1}^\times$, is back-propagated to $I_1^-$ as $I_p^\times$ in one period. In period $t=2$, the tainted fraction in $I_1^-$ is $z_2 = \frac{(a+1)|I_{p-1}^\times|}{|I_1^-|}$. In the first phase of this period, the impossibility region will propagate to $I_{p-1}^+$ so that $I_{p-1}^+$ will be partitioned into three parts with the ratio: $|v| : |I_{p-1}^\times| : |w|$, where the middle part is the new impossibility zone $I_{2p-1}^\times$. This ratio is precisely the same as the partitioning from period $t=1$. Since $\frac{|w|}{|v|} > a^{p-1}(a+1)$ and $|I_{2p-1}^\times| = (a+1)a^{p-1}$, we have $I_{2p-1}^\times \cap w \neq \emptyset$. That is, part of $w$ will be tainted with impossibility, which will then back-propagate to $I_1^-$ at period $t=3$. Let $B_1 = I_{2p-1}^\times \cap w$ be that which will back-propagate to $I_1^-$ in the next period. Then $z_3 = (a+1)\frac{|B_1|+|I_{p-1}^\times|}{|I_1^-|}$. In the first phase of period 3, there will be a new impossibility zone $B_2$ in $v$ to back-propagate toward $I_1^-$ at next period. In addition, $|B_2| = (a+1)a^p|B_1|$. Then $z_4 = (a+1)\frac{|B_2|+|B_1|+|I_{p-1}^\times|}{|I_1^-|}$. In the first phase of period 4, there will be a new impossibility zone $B_3$ in $w$, where $|B_3| = (a+1)a^{p-1}|B_2|$. Then $z_5 = (a+1)\frac{|B_3|+|B_2|+|B_1|+|I_{p-1}^\times|}{|I_1^-|}$. Following this pattern, at period $t$, the additional impossibility zone to be back-propagated to $I_1^-$ is $B_{t-2}$, where $|B_{t-2}| \geq a^{p-1}(a+1)|B_{t-3}|$. All the impossibility zones in $I_1^-$ form a contiguous interval. Therefore, at some finite $\kappa$, the impossibility zone will eventually taint the whole interval $[r_p, r_t]$, and $z_\kappa$ will reach 1. The case with $\frac{|w|}{|v|} < \frac{1}{a^p(a+1)}$ has a similar proof showing that there are no **PP** and **TT** strategies for $P_1$.

If $\frac{1}{a^p(a+1)} \leq \frac{|w|}{|v|} \leq a^{p-1}(a+1)$, we then have $I_{2p-1}^\times \cap w = \emptyset$ and $I_{2p-1}^\times$ is the last non-empty impossibility zone. The fraction tainted stays as $z_2 = \frac{(a+1)|I_{p-1}^\times|}{|I_1^-|} < 1$. Thus, for the initial I-states that do not belong to the impossibility zone, there is a privacy-preserving tracking strategy. Let $L' = \bigcup_{0 \leq j \leq 2p-1} I_j^\times$. The privacy-preserving tracking strategy for $[r_p, r_t] \setminus L'$ is as follows: take action $\oplus$ in $[r_p, r_t] \setminus (L' \cup I_1^-)$, and action $\ominus$ in $I_1^-$. The resulting I-state will always stay within $[r_p, r_t] \setminus L'$. ■

Lemma 12 says that the boundary problem in Figure 12d contains both initial I-state dependent cases and also instances which have no privacy-preserving tracking strategy.

### C. Interpretation and further observations

The preceding analysis of the boundary problem completes our characterization of all panda tracking problems. The earlier examination of Figure 6 can now be supplemented with the observation that the region marked in pink, already (from Lemma 7) known not to contain strategies for all initial I-states, actually consists of a mixture of colors. If we imagine an axis with the initial I-state sizes normal to the page, then the pink region is partly gray (Lemmas 9 and 11) and partly a mixture of green and gray (Lemmas 10 and 12), with the latter mixture having at least some gray on every line segment departing the page.

This analysis also illuminates some new possibilities for the problem. Under the assumption of a strong poacher who knows all the information obtained by the robot immediately, there is no hope to save the panda from the situations that are not privacy-preservable or target trackable. But if we relax this admittedly very strong model of the adversary, some additional versions of the problem can be salvaged.

Assume that there is some maximum amount of time, $\tau$, that the poacher is willing to spend hunting before giving up. Then the impossibility zones $I_t^\times, t \geq \tau$ become safe. Some problems, previously without privacy-preserving strategies, will become I-state dependent, and the I-state dependent problems will have more initial I-states that have privacy-preserving strategies.

Another model to consider is the case where the robot can purposefully forget some limited initial information, say, replacing $\eta_0$ with $\eta_0'$. Then the poacher can have all the information available, as before, excepting $\eta_0$. In this model all the I-state dependent problems can become privacy-preservable and target trackable by simply disguising the impossible initial I-states, which is easily achieved by expanding the I-state until it becomes a privacy-preservable and target trackable one.

## VI. BEYOND ONE-DIMENSIONAL TRACKING

The inspiration for this work was the 2-dimensional case. This section lifts the impossibility result to higher dimensions.

### A. Mapping from high dimension to one dimension

In the $n$-dimensional privacy-preserving tracking problem, the state for the panda becomes a point in $\mathbb{R}^n$. The panda can move with a maximum distance of $\frac{\delta}{2}$ in any direction in $\mathbb{R}^n$ within a single time-step, so that the panda's actions fill an $n$-dimensional ball. The privacy and tracking bound are also generalized from an interval of size $r_p$ and $r_t$, to an $n$-dimensional ball of diameter $r_p$ and $r_t$ respectively. That is, the I-state should contain a ball of diameter $r_p$ and be contained in a ball of diameter $r_t$, so as to achieve privacy-preserving tracking. The robot inhabits the $n$-dimensional space as well and attention must be paid to its orientation.

It is unclear what would form the appropriate higher dimensional analogue of parameter $c$, so we only consider $n$-dimensional tracking problems for robots equipped with a generalization of the original quadrant sensor. The sensor's orientation is determined by that of the robot and it indicates which of the $2^n$ possible orthogonal cells the panda might be in. Adopting notation and definitions analogous to those earlier, we use a tuple for $n$-dimensional tracking problems—a subscript makes the intended dimensionality clear.

The following lemma shows that there is a mapping which preserves the tracking property from $n$-dimensional problem to 1-dimensional problems.

**Lemma 13.** *Given some 1-dim. panda tracking problem $P_1 = (\eta_0, r_p, r_t, \delta, n)$, there exists an $n$-dim. panda tracking problem $P_n = (\theta_0, r_p, r_t, \delta, 1^n)$ where, if $\mathbf{TT}(P_n)$, then $\mathbf{TT}(P_1)$.*

*Proof:*
The approach to this proof has elements of a strategy stealing argument and simulation of one system by another. The robot faced with a 1-dim. problem constructs an $n$-dim. problem and uses the (hypothetical) strategy for this latter problem to select actions. The crux of the proof is that the 1-dim. robot can report back observations that are apposite for the $n$-dim. case. Figure 13, below, gives a visual overview.

For some $P_1 = (\eta_0, r_p, r_t, \delta, c)$, with $c = n$, we construct $P_n = (\theta_0, r_p, r_t, \delta, 1^n)$ as follows. Without sacrifice of generality, assume that in $P_1$ the initial I-state $\eta_0 = \{x \mid \eta_0^{\min} \leq x \leq \eta_0^{\max}\}$ is centered at the origin, so $\eta_0^{\min} = -\eta_0^{\max}$. (This simplifies the argument and a suitable translation of coordinate system rectifies the situation otherwise.) Then we choose $\theta_0$ as the closed ball at the origin with radius $\eta_0^{\max}$.

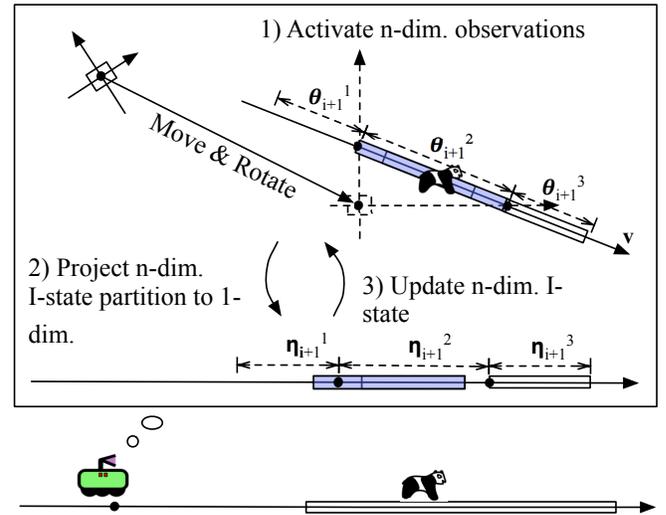

Fig. 13: Constructing a 1-dim. strategy $\pi_1$ from some $n$-dim. strategy $\pi_n$.

We show how, given some $\pi_n$ on $P_n = (\theta_0, r_p, r_t, \delta, 1^n)$, we can use it to define a $\pi_1$ for use by the 1-dim. robot. The robot forms $\theta_0$ and also has $\eta_0$. It picks an arbitrary unit-length vector $\hat{\mathbf{v}} = v_1 \mathbf{e_1} + v_2 \mathbf{e_2} + \cdots + v_n \mathbf{e_n}$, unknown to the source of $\pi_n$, which is the subspace that the 1-dim. panda lives in. For subsequent steps, the robot maintains $\theta_1^-, \theta_1, \theta_2^-, \theta_2, \ldots, \theta_k^-, \theta_k, \theta_{k+1}^-, \ldots$ along with the I-states in the original 1-dim. problem $\eta_1^-, \eta_1, \eta_2^-, \eta_2, \ldots, \eta_k^-, \eta_k, \eta_{k+1}^- \ldots$ For any step $k$, the $\eta_k$ can be seen as measured along $\hat{\mathbf{v}}$ within the higher dimensional space. Given $\theta_{k-1}$, $\theta_k^-$ is constructed using Minkowski sum operations as before, though now in higher dimension. Given $\theta_k^-$, strategy $\pi_n$ determines a new pose for the $n$-dim. robot and, on the basis of this location and orientation, the $n$ sensing planes slice through $\theta_k^-$. Though the planes demarcate $2^n$ cells, the line along $\hat{\mathbf{v}}$ is cut into no more than $n+1$ pieces as the line can pierce each plane at most once (with any planes containing $\hat{\mathbf{v}}$ being ignored). Since the 1-dim. robot has $c = n$, it picks the $u_1, \ldots, u_n$ by measuring the locations that the sensing planes intersect the line $\mathbf{x} = \alpha\hat{\mathbf{v}}, \alpha \in \mathbb{R}$. (If fewer than $n$ intersections occur, owing to planes containing the line, the extra $u_i$'s are simply placed outside the range of the I-state.) After the 1-dim. panda's location is determined, the appropriate orthogonal cell is reported as the $n$-dim. observation, and $\theta_k^-$ leads to $\theta_k$ via the intersection operation. This process comprises $\pi_1$. It continues indefinitely because $\theta_k$ must always entirely contain $\eta_k$ along the line through $\hat{\mathbf{v}}$ because, after all, a cantankerous $n$-dim. panda is free to choose to always limit its movements to that line.

If $\pi_n$ is **TT**, then so too is the resulting strategy $\pi_1$ since the transformation relating $\theta_k \cap \{\alpha\hat{\mathbf{v}} : \alpha \in \mathbb{R}\}$ with $\eta_k$ preserves length and, thus, $\theta_k$ fitting within a ball of diameter $r_t$ implies that $|\eta_k| < r_t$. ∎

### B. Impossibility in high-dimensional privacy-preserving tracking

Now we are ready to connect the pieces together for the main result:

**Theorem 3.** *(Impossibility) It is not possible to achieve privacy-preserving panda tracking in $n$ dimensions for every problem with $r_p < r_t$.*

*Proof:* To extend the lemmas that have shown this result for $n = 1$ to cases for $n > 1$, suppose such a solution existed for $P_n = (\theta_0, r_p, r_t, \delta, 1^n)$. Then according to Lemma 13, every 1-dim. panda tracking problem $P_1 = (\eta_0, r_p, r_t, \delta, n)$ is **TT**, since they can be mapped to an $n$-dim. **TT** panda tracking problem. But this contradicts the non-**TT** instances in Lemma 4, so no such strategy can exist for every non-trivial problem ($r_p < r_t$) in two, three, and higher dimensions. ∎

Theorem 3 (via Lemma 13) relates an $n$-dim. robot equipped with a generalized quadrant sensor (that is what the $1^n$ denotes) to a 1-dim. robot with a sensor capability value of $n$. The $n$-dim. sensor has $2^n$ separate output classes (or preimages), yet the 1-dim. sensor has only $n+1$. It appears, at first sight, that the reduction is to a *less* capable sensor, which seems paradoxical as some information goes missing. But the 1-dim. robot selects the boundaries of the preimages in a far more flexible way than any higher dimensional robot would acheive with actions that move (and reorient) the quadrant sensor's origin.

## VII. CONCLUSION AND FUTURE WORK

In this paper we have reexamined the panda tracking scenario introduced by O'Kane (2008), focusing on how various parameters specifying a problem instance, including the capabilities of the robot and the panda, affect the existence of solutions. Our approach has been to study nontrivial instances of the problem in one dimension. This allows for an analysis of strategies by examining whether the sensing operations involved at each step increase or decrease the degree of uncertainty in a directly quantifiable way. Only if this uncertainty can be precisely controlled forever, can we deem the problem instance solved. We use a particular set of sensing choices, basically division of the region evenly, which we think of as a *split* operation. This operation is useful because the worst resulting I-state in other choices, those that are not evenly split, is weaker (in terms of satisfying the tracking and privacy constraints) than even divisions are. Thus, the split operation acts as a kind of basis: if the problem has no solution with these choices, the panda cannot be tracked with other choices either.

In examining the space of tracking and privacy stipulations, the existence of strategies is shown to be a function of the robot's initial belief and panda's movement. There exist regions without solution, where it is impossible for the robot to actively track the panda as well as protect its privacy for certain nontrivial tracking and privacy bounds. Additionally, we have uncovered regions where solution feasibility is sensitive to the robot's initial belief, which we have called I-state dependent cases (or conditions). The simple one-dimensional setting also permits exploration of how circumstances change as the robot's sensing power increases. Perhaps surprisingly, the number of these I-state dependent strategy conditions does not decrease as the robot's sensing becomes more powerful. Finally, we connect the impossibility result back to O'Kane's setting by mapping between high-dimensional and one-dimensional versions, proving that the 2D planar panda tracking problem does not have any privacy-preserving tracking strategy for every non-trivial tracking and privacy stipulation.

The results presented in this paper reveal some properties of a particular—and it must be said somewhat narrow—instance of an informationally constrained problem. Future work that explores what sensor properties permit such a task to be achieved, perhaps identifying categories of sensor families that suffice more broadly, would certainly be useful. Also, probabilistic models impose different belief representations and observation functions and it is worth exploring analogous notions in those settings. Another thread is to weaken the poacher's capability, allowing the privacy bound to be relaxed somewhat. For example, in considering some latency in the poacher's reaction to information that has been leaked, it may be acceptable for the privacy bound to be relaxed so that it needs not apply from every time-step to the next.

ACKNOWLEDGMENTS

This work was supported by the NSF through awards IIS-1527436 and IIS-1453652. The pirate image in Figure 1 is adapted from the work of J.J. and Gustavb under a Creative Commons license.